\documentclass{article} 
\usepackage{iclr2027_conference,times}

\usepackage{amsmath,amsfonts,bm}

\def\eqref#1{equation~\ref{#1}}

\def\1{\bm{1}}

\DeclareMathAlphabet{\mathsfit}{\encodingdefault}{\sfdefault}{m}{sl}
\SetMathAlphabet{\mathsfit}{bold}{\encodingdefault}{\sfdefault}{bx}{n}

\usepackage{hyperref}
\usepackage{url}
\usepackage{graphicx}
\usepackage{subcaption}
\usepackage{algorithm}
\usepackage{wrapfig}
\usepackage{algpseudocode}
\usepackage{booktabs}
\usepackage[dvipsnames]{xcolor}
\usepackage{amsthm}
\newtheorem{informaltheorem}{Informal Theorem}
\usepackage{amsthm}

\usepackage{multirow}
\newcommand{\model}{LinSlot}

\title{\model: Exploiting Linear Representation hypothesis for unsupervised attribute discovery from slot based object representation}
\iclrfinalcopy 
\author{Sanket Gandhi$^{1}$, Utkarsh Giri$^{1}$, Varun Subramaniam$^{2}$, Rohan Paul$^{1,2}$ \& Parag Singla$^{1,2}$ \\
$^{1}$Yardi School of AI, IIT Delhi \quad $^{2}$Department of Computer Science and Engineering, IIT Delhi \\
Correspondence to: \texttt{sanket.gandhi@scai.iitd.ac.in}
}

\begin{document}

\maketitle
\maketitle
\begin{abstract}
    This paper studies the problem of learning disentangled representations of objects and their attributes from raw, unstructured image data. Slot-based methods have shown considerable success in unsupervised learning of object representations from images. Block-slot attention-based methods extend this framework to attribute representations by assuming a uniform factorization of object representations into attributes, which may be suboptimal and consequently limit the quality of the learned representations. We therefore investigate a framework for jointly discovering object and attribute representations. Our key contribution is leveraging the Linear Representation Hypothesis (LRH), which postulates that composable concepts can be represented as linearly additive subspaces in slot representations. Based on this insight, we propose a probabilistic model connecting images, slots (objects), and blocks (attributes). We present an architecture that leverages block attention to connect attribute representations to slots and incorporates LRH in both object and attribute representation spaces. This architecture effectively optimizes the Evidence Lower Bound (ELBO) of the proposed graphical model. Our experiments demonstrate (i) effective discovery of disentangled object and attribute representations, (ii) empirical evidence for LRH in  slot space, and (iii) the ability to perform image editing owing to the disentangled and interpretable nature of the learned representations. Our experiments on multiple datasets demonstrate improvements in DCI scores over state-of-the-art methods.

\end{abstract}
\section{Introduction}
A core aspect of the human intelligence is its ability to form disentangled representations of the world \citep{fodor,symbolism, mental_program}, without having any access to supervised data. Enabling deep learning models with such representations has been a long-standing goal and is desirable for a number of reasons, i.e., having properties such as interpretability, compositional generalization, and sample efficiency \citep{representationsurvey, binding, buildingmachine, inductivebiases}. In this work, we are specifically interested in the task of joint disentanglement of objects and their attributes (such as color, shape, size etc.) in static images without any supervision. Extracting such information automatically from images, in an unsupervised manner, will go a long way in building models, that have above desired properties of generalizability, sample efficiency and robustness to noise. Recently proposed \textit{slot attention} based methods have shown promising results on learning \textit{object-centric} disentangled representation in images\citep{slot_attention,slotadapt25,ctrlo25}, and these methods have been effectively applied for downstream tasks such as image segmentation, compositional generation and VQA. Extending this work, block-slot attention~\citep{nsb,ncb24} based methods partition the learned object representations (slots) into attributes (factors) (e.g., color, shape) yielding a fine grained and interpretable understanding of the image generation process. 

Despite this progress, a fundamental question remains yet to be answered: is there a theoretical framework, in the form of a generative graphical model, which can guide the joint discovery of objects and their attributes in images in an unsupervised manner and provide a guarantee on the identifiability of latent factors under suitable assumptions? If so, what are these assumptions under which such identifiability of factors (up to a permutation) holds, and how can we design an encoder-decoder architecture based on the assumptions of our model, to effectively discover the objects, and their latent factors from a given set of images.

Our solution to this problem is motivated by the linear representation hypothesis which has been used extensively for better understanding of latent factors in textual domains~\cite{park24linear}. There have been some recent attempts to extend this work to the visual domains ~\citep{sae_gao, sae_dino}, and that becomes our starting point. Specifically, we make an assumption that the object can be represented as a stochastic linear combination of its attribute values sampled from a given prior. Under this assumption, our graphical model defines a three-step sampling process: (1) sampling a concept index (e.g., red, blue, or green) and its corresponding value (e.g., the degree of redness) from a prior distribution; (2) constructing each object representation as a linear function of the sampled factor values; and (3) generating the image through a learned function of the resulting object representations.
 This graphical model enables derivation of an ELBO (evidence-lower-bound) formulation for optimizing the likelihood. 

Correspondingly, our architecture, for disentanglement, involves a two-step encoder, going from an image to corresponding slots using existing slot-attention based methods~\citep{slot_attention} in the outer loop, which formulates this as a problem of competing attention over image patches (2) followed by block discovery module, which first projects the slot to a set of latent dimensions, over which blocks attend via competing attention, for the discovery of latent factors. The factors are poised to be linear combination of underlying (learnable) concepts, in line with earlier work~\citep{nsb}, albeit with strong justification arising from our graphical model, for using the linear combination, as a way of imposing mixture of Gaussian prior. Decoding works in reverse, by first constructing the slots from the discovered factors, followed by construction of the image from the decoded slot. We experiment with two variants of our approach, one using a decoder working directly on the slot representation, and the other one, where the internal structure of each slot, in terms of its constituent blocks, is exposed to the decoder. Entire architecture is trained in an end to end manner, with a curriculum, for the joint discovery of objects (slots) and their attributes (factors).

We first show experimentally the validation of the linear structure in the slot space across the slots methods and datasets. We demonstrate the object manipulation via simple slot arithmetic of addition and subtraction. We perform extensive evaluation over three benchmark datasets to evaluate the efficacy of our approach both for object discovery as well as factor discovery. We find that our two stage pipeline not only results in improved factor discovery, it also improves the object discovery, compared to existing baselines, both in quantitative as well as qualitative evaluation. We also perform ablations demonstrating the contribution of various components our two-stage pipeline towards the efficacy of object and attribute discovery. We plan to release our code publicly on
acceptance.

\section{Related Work}
\textbf{Object-Centric Learning}: Unsupervised object discovery methods aim at learning disentangled object representations for scenes without supervision. Earlier methods used generative modeling of the scene and imposed the object level structure in the latents \citep{air, squair, scalor, gswm, stove, rsqair, supair}. These \textit{spatial attention methods} struggle to find occluded and non-local objects. \textit{Scene mixture methods} models scenes as composition of object images \cite{nim, iodine, monet, gensis, gensis2} - however inference is computationally challenging. \emph{Slot attention} \cite{slot_attention} leverages competitive spatial attention over the image to allow \emph{slots} to bind to objects in an image. Others \citep{savi, savi++, slotformer} model temporal dynamics and discover objects from images. Recent efforts extend the approach to realistic scenes by incorporating more expressive decoders \cite{slate, steve, lsd, slotdiff} as well as using pre-trained image features \citep{dino} combined with a diffusion decoder \cite{dinosaur, videosaur, glass,slotadapt, metaslot, coda}. 

\textbf{Disentangled representation learning}: Attribute discovery aims at learning independent factored representations such as colour, shape, texture etc. that explain generation of objects observed in images. 
Earliest approaches train an VAE \citep{vae} based methods \citep{bvae, fvae, tbvae} but suffer from identifiability issues \citep{challenging, ica_vae} necessitating the need for inductive biases to aid disentanglement. However with weak supervision it is guaranteed to recover attributes \citep{weaklywithgrt, weaklywithsp, weaklywthcom}. Recently diffusion and transformer based methods were also proposed \cite{vct, disdiff}. Enabling compositionality of representations also enabled disentanglement \cite{compbias}.

\textbf{Linear Representation Hypothesis}: Presence of the linear structure in the textual domain well established \citep{glove, kiho1, kiho2, llm_lrh} even in LLMs. In vision domain  \cite{vlrh1, vlrh2}. However recovering such representations without supervision is difficult \cite{challenging}. 
Existing work uses Sparse Autoencoders (SAEs) for recovering the \textit{concepts} from latent representation which is additive in concepts\citep{sae_gao, sae_dino}. However  block have to identify all the values of the specific factor and SAE atoms fail to capture such representations \citep{sae_manifold}. One option is to enable grouping of SAE atoms \citep{block_sae} but it assumes the presence of the attributes is sparse which is not true for our proposed generative model.

\section{Background: Slot Attention}

Mathematically, given an input image $X \in \mathbb{R}^{H \times W \times 3}$ it is first passed through an encoder typically implemented as CNN to get the latent representation of the image, as a vector as $\mathbf{E} \in \mathbb{R}^{N_{e} \times d_{input}}$, where $N_{e}$ represents the number of spatial features, and $d_{input}$ represents their dimension. Following this, the each element in the slots set $\mathbf{S} \in \mathbb{R}^{N_s \times d_{slot}}$ is initialized by sampling from a normal distribution $\mathcal{N}(\mu, \Sigma)$ with learnable parameters. These slots then attend to the input image features with learnable $k$, $q$ and $v$ projection functions (matrices) to get the attention matrix $A$ as: 
\begin{equation} 
\label{eqn:slot_attn}
A = \texttt{Softmax} \left ( \frac{q(\mathbf{S})k(\mathbf{E})^T}{\sqrt{d_{slot}}}, \texttt{axis='slots'} \right ) 
\end{equation}
Additionally, the attention matrix $A$ is further weight normalized along the 'axis' dimension to ensure competitive nature of the attention. The update vector $U = A*v(\mathbf{E})$, resulting from the attention mechanism, is then fed to a recurrent neural network as input and previous slots as states. The output of the recurrent neural network is passed through an MLP to get the updated slot representation. This process is repeated for fixed number of iterations to get the final slot representation. Refer to Algorithm 1 in ~\cite{slot_attention} for a detailed pseudocode.

\section{\model}

In this section, we present our framework \model{} for unsupervised
factor representation learning from \textit{slots}. We first
formulate a generative model of images that exploits the
\textit{linearity} of the slot representation space
(Section~\ref{subsec:generative_model}). We then establish
identifiability of the proposed generative formulation under weak
assumptions (Section~\ref{subsec:identifiability}). Next, we describe
a deterministic inference procedure motivated by this formulation, which first
discovers object slots using Slot Attention and then infers
attribute representations using our Block Attention mechanism
(Section~\ref{subsec:inference}). Finally, we describe the training
objectives motivated by the resulting ELBO and our curriculum-based
training procedure (Section~\ref{subsec:learning}).



\begin{figure}[t]
    \centering

    \begin{subfigure}[t]{0.30\linewidth}
        \centering
        \includegraphics[width=\linewidth]{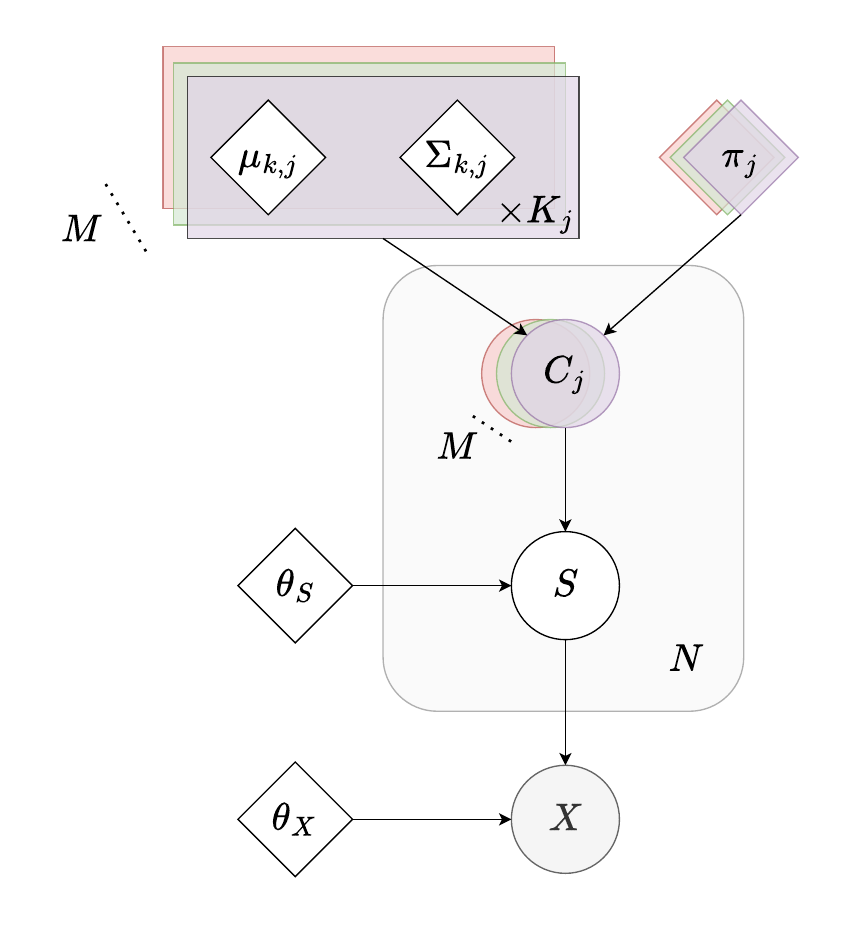}
        \caption{Generative model.}
        \label{fig:generative_model}
    \end{subfigure}
    \hspace{0.07\linewidth}
  \begin{subfigure}[t]{0.40\linewidth}
        \centering
        \includegraphics[width=\linewidth]{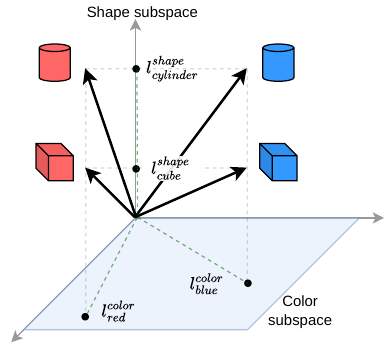}
        \caption{Linear representation of slots.}
        \label{fig:lin_slot}
    \end{subfigure}
    \caption{Overview of the proposed representation and generative model.}
    \label{fig:overview}
\end{figure}

\subsection{generative model} \label{subsec:generative_model}
In this section, we describe the generative model of the images considered in our work. We assume that each image consists of a set of objects, where each object is characterized by a set of factors of variation, such as color, shape, and size. Each factor is characterized by a fixed set of concepts, and each factor is associated with a particular concept from this set. For example, red, blue, and green are concepts for the color factor. The generative process proceeds in three steps: (1) for each factor of each object, a concept index is first sampled, after which the corresponding concept is sampled from its own prior and assigned to that factor. We assume the concept prior to be a normal distribution, which results in a GMM for the distribution of each factor; (2) the representation of each object is sampled conditioned on its factor values, and specifically, is given by the linear sum of the factor representations; and (3) the image is generated conditioned on the representations of all objects. Fig.~\ref{fig:generative_model} illustrates this generative process.

Our problem statement then is: given a set of training images, a family of prior distribution $p$ over concepts, an unknown function $f$ mapping the sampled concept values to the corresponding object representation, and another function $g$ mapping the set of object representations to the image representation, under what conditions, can we uniquely discover the (a) parameters of the concept prior (b) values of the underlying concepts upto a permutation, (c) inverse of the functions $f$, given fixed $g$? This leads to the following formulation and the corresponding identifiability result.

Specifically, assume that the image $X \in \mathbb{R}^{H \times W \times 3}$ is generated by set of $N$ latent object representations referred as \textit{slots} decoded by ${\textbf{S}} = \{s_1,...,s_N\}$. Each slot $s_i \in \mathbb{R}^{d_{\texttt{slot}}}$ itself is independently generated by $M$ latent factor (attribute) representations referred as \textit{blocks} denoted by the ordered set $\textbf{c}_i = \{c_{1,i},..,c_{M,i}\}$. Each $j^{th}$ block $c_{j,i} \in \mathbb{R}^{d_{\texttt{block}}}$ is sampled independently from it's own specific prior which is shared across all the objects. Let $\textbf{C} = \{\textbf{c}_1,..,\textbf{c}_N\}$. The generative process can be expressed by following joint distribution:
\begin{equation}
    p_{\theta}(X,\mathbf{S},\mathbf{C}) = p_{\theta_X}(X|\textbf{S})\prod_{i=1}^N \left ( p_{\theta_S}(s_i | \textbf{c}_i) \prod_{j=1}^M p_{\theta_{C_j}}(c_{j,i}) \right ) 
\end{equation}

Where $\theta = \{\theta_X, \theta_S, \{\theta_{C_j}\}_{j=1}^M \}$ are the parameters of the distribution. We further assume the prior on the blocks to be a Gaussian mixture model (GMM) which are specific to the blocks.  
\begin{equation}
    c_{j,i} \sim p_{\theta_{C_j}}(C_j) := \sum_{k=1}^{K_j} \pi_{k,j}\mathcal{N}(\mu_{k,j}, \Sigma_{k,j} )
\end{equation}
\underline{\textbf{Linear Structure of Slots}}\label{note:linear_structure}:
We also restrict the slot representations to be \textit{linearly additive} in terms of some transformed block representations. That is 
\begin{equation}
    s_{i} \sim p_{\theta_S}(S|\textbf{c}_{i}) := \mathcal{N}(\mu(\textbf{c}_i), \Sigma_S) \ \ \ \\ \ where \ \ \ \ \ \mu(\textbf{c}_i) = \sum_{j=1}^M f_{j}(c_{j,i}) + b
\end{equation}
Such additive structure implies the presence of the global \textit{delta directions} in slots space which represents a direction from one factor value to another value. For example in Fig. \ref{fig:lin_slot} the delta direction of the blue to red color and cube to cylinder shape denoted as $\delta_{color}[red \to blue]$.
These \textit{delta directions} can be used to manipulate factors in the slot representation by simply vector addition. For example adding the delta direction   $\delta_{color}[red \to blue]$ to any red object slot should change the object color to blue. 

\subsection{identifiability} 

\begin{informaltheorem}[Identifiability of the Blocks ]
Under the proposed generative model and the assumptions that  $\{f_j\}_{j=1}^M$ are  piecewise affine,  $Md_{\texttt{block}} \leq d_{\texttt{slot}}$ and $K_{i} \neq K_{u}K_{w}$ for any triplets $i,u,w \in {1,..,M}$, then the blocks are identifiable upto translation and permutation given slots are fixed. 
\end{informaltheorem}

 The proof is provided in the \ref{appendix:proof_of_identifiablity} which follows from the \cite{identifiability_gmm} where we first show that the concatenated blocks are also from GMM. Then the number of 
\label{subsec:identifiability}
\subsection{inference} \label{subsec:inference}
We approximate the true posterior $p_{\theta}(\textbf{S}, \textbf{C} |X)$ with variational approximation $q_{\phi}(\textbf{S}, \textbf{C} | X)$ due to intractability of the former. We further factorize the variational posterior distribution as \begin{equation} \label{eq:inference_dist}
    q_{\phi}(\textbf{S}, \textbf{C} | X) = q_{\phi_S}(\textbf{S} | X)\prod_{i=1}^N q_{\phi_C}(\textbf{c}_i | s_i)
    \end{equation}
One benefit of having such inference modeling is that the slot posterior $q_{\phi}(\textbf{S} | X)$ can be implemented as any slot based method and could also be pretrained. The block discovery method is not tied to the slot discovery as done in the block-slot attention which makes the optimization simpler. However inferring the blocks from slots is not trivial and we discuss it in following section.

The factorization in Eq.~\ref{eq:inference_dist} suggests a sequential inference
procedure: given an image $X$, we first infer the object slots
$\mathbf{S}$ using a slot-based object discovery method, and then
infer the attribute blocks $\mathbf{C}$ conditioned on the inferred
slots. In our implementation, we use Slot Attention for the first
stage and introduce Block Attention for the second stage:
\begin{equation}
X \xrightarrow{\text{Slot Attention}} \mathbf{S}
\xrightarrow{\text{Block Attention}} \mathbf{C}.
\end{equation}

The block inference module should therefore infer the blocks using
the slots as input. Note that the slot representation is additive in
terms of transformed block representations.

\subsubsection{Block inference using block attention} \label{sec:block_attention}
\begin{wrapfigure}{r}{0.50\columnwidth}
    \vspace{-10pt}
    \centering
    \caption{Block Attention}
    \label{alg:block_attention}

    {\footnotesize
    \begin{algorithmic}[1]
    \State Input: $s_i$
    \State Initialize $\bar{\textbf{c}}_{i} \sim \mathcal{N}(\mu_C, \Sigma_C)$
        
        \State $c_{j,i} = \bar{c}_{j,i}$
            $\forall j \in \{1,\ldots,M\}$

        \State $v_{k,i} = (s_i^\top u_k)u_k$
            $\forall k \in \{1,\ldots,N_p\}$

        \For{$t = 1$ to $T_b$}

            \State $c_{j,i} =
            \texttt{LN}(c_{j,i})$
            $\forall j \in \{1,\ldots,M\}$
            \color{blue}\State $A_i =
            \texttt{Softmax}\left(
            \frac{k(\mathbf{V}_i)q(\mathbf{c}_i)^\top}
            {\sqrt{d_{\mathrm{block}}}}
            \right)$

            \State $A_i =
            A_i / A_i.\texttt{sum}(\texttt{d='input'})$

            \State $U_i = A_i^Tv(\mathbf{V}_i)$

            \color{black}\For{$j \in \{1,\ldots,M\}$}

                \color{red}\State $u_{j,i} =
                \texttt{GRU}_j(
                \texttt{st}=c_{j,i},
                \texttt{in}=u_{j,i})$
                \State $u_{j,i} =
                u_{j,i} + \texttt{MLP}_j(u_{j,i})$
                \color{OliveGreen}\State $c_{j,i} =
                \texttt{Softmax}\left(
                \frac{u_{j,i}\mathcal{M}_j^\top}
                {\sqrt{d_{\mathrm{block}}}}
                \right)\mathbf{\mathcal{M}}_j$\color{black}
            \EndFor
    \EndFor
    \State \textbf{return} $\mathbf{c}_i$

    \end{algorithmic}
    }

    \vspace{-8pt}
\end{wrapfigure}

 We propose a \textit{iterative refinement} based inference algorithm for inferring blocks from slots. Blocks are refined with the attention between  projections of the slots along fixed learnable directions and the blocks inferred so far. The attention is normalized such that the blocks compete to attend for a particular projection of the slot. 

Specifically, 
let $u_1,,..u_{N_p}$ denote the $N_p$ learnable vectors representing the latent space of projections where each $u_k \in \mathbb{R}^{d_{slot}}$. Let the projections of the slot $s_i$ onto this space be denoted as $\mathbf{V}_i^T = [v_{1,i},...,v_{N_p,i}]$ where $v_{k,i} = (s_i^Tu_k)u_k$. Given these projection vectors represented in the latent space, we infer the blocks corresponding to a slot $s_i$ by iteratively refining initial block set $\bar{\mathbf{c_{i}}}$ with the attention on to these projection vectors. The attention is made competitive by normalizing along block axis (Line 7, Algorithm~\ref{alg:block_attention}). In line 8, we normalize the attention matrix again along input (project vector) axis, normalization across 'input' (projection vectors) dimension. $U_i$ (line 9) denotes the update vector computed as a result of the attention mechanism. 

The updates corresponding to particular blocks $c_{j,i}$ denotes as $u_{j,i} \in \mathbb{R}^{d_{block}}$ are used to refine the blocks from previous iterations with the GRU and MLP shown in the algorithm. Finally at the end of each refinement step, the blocks are also replaced as linear sum of the prototypes similar to block slot attention~\cite{nsb} (line 13).


\captionsetup{skip=0pt}
\begin{figure*}[t] 
    \centering
\includegraphics[width=\textwidth]{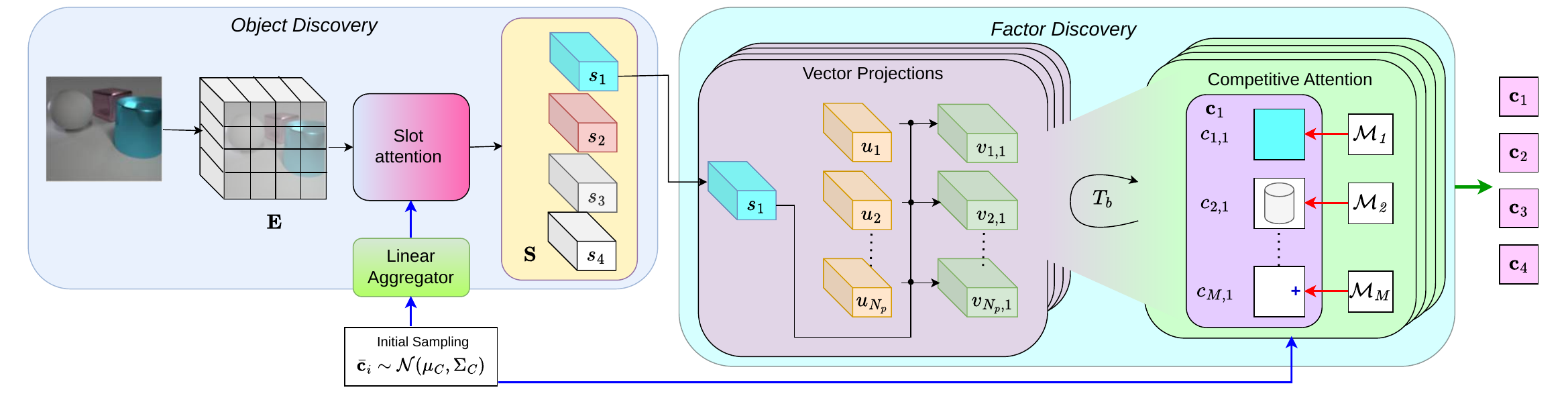} 
\caption{\textbf{Overview of the inference:} The image is encoded into spatial features $\mathbf{E}$, from which slots $\mathbf{S}$ are inferred via slot attention. For each slot, projected features $\mathbf{v}$ are used to iteratively refine $M$ attribute blocks through competitive attention, with each block constrained to a linear combination of shared concept vectors. The resulting blocks serve as disentangled attribute representations.
}
    \label{fig:wide_diagram}
\end{figure*}

\subsection{Learning and Implementation} \label{subsec:learning}

The parameters $\theta$ and $\phi$ can be learned by maximizing the evidence lower bound (ELBO) of the data log-likelihood. For the proposed generative model, the ELBO is given by:
For the full derivation, please refer to Appendix~\ref{sec:elbo_derivation}.
\begin{equation}
\begin{aligned}
\log p_{\theta}(X) \geq\;&
\underbrace{\mathbb{E}_{q_{\phi_S}(\mathbf{S}|X)}
\left[\log p_{\theta_X}(X|\mathbf{S})\right]}_{\text{Reconstruction}}
-
\underbrace{D_{KL}\!\left(q_{\phi_S}(\mathbf{S}|X)\middle\|
p_{\theta_S}(\mathbf{S}|\mathbf{C})\right)}_{\color{red}\text{Slot regularizer}} \\
&-
\underbrace{
\sum_{i=1}^N
\mathbb{E}_{q_{\phi_S}(\mathbf{S}|X)}
\left[
D_{KL}\!\left(q_{\phi_C}(\mathbf{c}_i|s_i)\middle\|
p_{\theta_C}(\mathbf{c}_i)\right)
\right]}_{\color{blue}\text{Block regularizer}}.
\end{aligned}
\end{equation}
The first term corresponds to the reconstruction objective, which measures the reconstruction error between the input images and the images reconstructed from the inferred slots. The second term is the slot regularizer, which encourages the inferred slots to be close to the linear combination of the transformed blocks. The third term is the block regularizer, which penalizes deviations of the inferred block distributions from their GMM priors.

In our implementation, we optimize the image reconstruction loss and the slot regularization loss explicitly, while the block regularizer is enforced implicitly through the prototype memory. Furthermore, when the mixture assignment probabilities are inferred using cosine similarity, the block projection in the prototype space (Line~13, Algorithm~\ref{alg:block_attention}) is equivalent to explicitly setting the block regularization loss to zero. See Appendix~\ref{sec:concept_projection_elbo} for the proof. Thus, we optimize the following loss:
\begin{equation}
    \mathcal{L}_{\mathrm{total}} =
    \mathcal{L}_{\mathrm{img-recon}} +
    \lambda \color{red}{\mathcal{L}_{\mathrm{slot-reg}}}\color{black}.
\end{equation}
We implement the slot-specific inference network $q_{\phi_S}$ using Slot Attention and the block-specific inference network $q_{\phi_C}$ using Block Attention (Section~\ref{sec:block_attention}). The first loss is the image reconstruction loss. Depending on the decoder architecture, this loss can be implemented as an MSE loss for CNN-based decoders, a dVAE prediction loss for transformer-based decoders, a denoising loss for diffusion-based decoders, or a feature reconstruction loss. The second loss is the slot regularization loss, defined as
$\mathcal{L}_{\mathrm{slot-reg}}
    = \sum_{i=1}^{N} \left\|s_i - \hat{s}_i\right\|$
where $s_i$ denotes the slots inferred by Slot Attention and $\hat{s}_i$ denotes the slots generated from the inferred blocks, given by
$
    \hat{s}_i = \sum_{j=1}^{M} f_j(c_{j,i}) + b.
$

We train the model in two phases. In the first phase, we train only the Slot Attention module using the image reconstruction loss. In the second phase, we introduce the block-specific modules and train the full model using the losses described above. For the slot-specific reconstruction loss, we found it useful to also use slots generated from the inferred blocks. Specifically, in each mini-batch, we use half of the slots inferred by Slot Attention and half of the slots generated from the inferred blocks. In 2nd phase, instead of sampling the initial slots from a normal distribution, we initialize them using linear combination of the transformed initial blocks. Specifically, the slot $s_i$ in Slot Attention is initialized as
$s_i = \sum_{j=1}^{m} f_j(\bar{c}_{j,i}).$
Line~2 of Alg.~\ref{alg:block_attention} for block initialization procedure.

\section{Experiments}

Our experiments try to answer the following questions: (1) Are slot representations linear in terms of the object factors? (2) Is the proposed attention-based inference method capable of recovering objects and factors of objects? (3) How much is the model robust to its training pipeline and its architecture decisions?
\subsection{Setup}
\underline{\textbf{Methods:}} We evaluated various slot based methods for the validating the linearity in their representations. This includes the 1. Slot attention \cite{slot_attention} which uses SBD decoder \cite{sbd}. 2. SLATE \cite{slate}  which uses the transformer based decoder. We found that modifying SLATE with difference slot initialization mechanism and a small module before decoder improves the SLATE performance. Please refer to Appendix \ref{appendix:Slate_modifications}. We name this baseline as SLATE+. 3. Slot-Diffusion \cite{slotdiff} which uses the diffusion decoder. 4. CODA-base \cite{coda} which uses the denoising diffusion loss from pretrained diffusion. We do not include the contrastive alignment loss from the original work. 

 For evaluating the unsupervised object and attribute discovery we use Slot attention, SLATE+ and a block-slot attention based baseline SysBinder \cite{nsb}. For our proposed attention based method \model \ we use SLATE+ as slot base which we train in the first phase. To see the effect of the attention based inference method alone we also include our attention based sequential inference method with \textit{block decoder} instead of slot decoder. We  name it as \model B. \model B is trained end to end without slot regularizer.


\underline{\textbf{Datasets:}}
We use three image benchmarks derived from \cite{Johnson2016CLEVRAD, clevrtex}, provided by \cite{nsb} for unsupervised object and attribute discovery. These datasets consist of the multi object images with $128 \times128$ resolution. The objects in these datasets are characterized by set of categorical and continuous attributes. CLEVR-Easy, CLEVR-Hard and CLEVR-Tex. Additionally for slot linearity validation we further use MOVi-C \cite{movi-c} which The frames from video as used as images.
We generate the paired version of the datasets where each datapoint is 4 images. The first two images are same except one object is different which have a single attribute different . The remaining two images are also same have all objects same except one object which are different in single attribute same as the attribute in first pair and the values of the changed attribute matches that from the first pair.  For further details please refer to \ref{appendix:datasets}

\underline{\textbf{Metrics:}} We evaluate the linearity of slot representation by cosine similarity between the two estimations of the \textit{delta directions} \ref{note:linear_structure}. That is subtraction of slot pair differing in only one attribute should be similar to the subtraction of different slot pair differing in same one attribute with similar values.  We evaluate the attribute disentanglement quality of representation using the standard DCI metric \cite{dci_paper}, which measures \underline{D}isentanglement, \underline{C}ompleteness, and \underline{I}nformativeness. Additionally, we report the Foreground Adjusted Rand Index (FG-ARI) to quantify object discovery. FG-ARI measures the clustering agreement between predicted slots and ground-truth masks exclusively on foreground objects, providing as assessment of segmentation quality.

\begin{figure*}[t]
    \centering
\includegraphics[width=\textwidth]{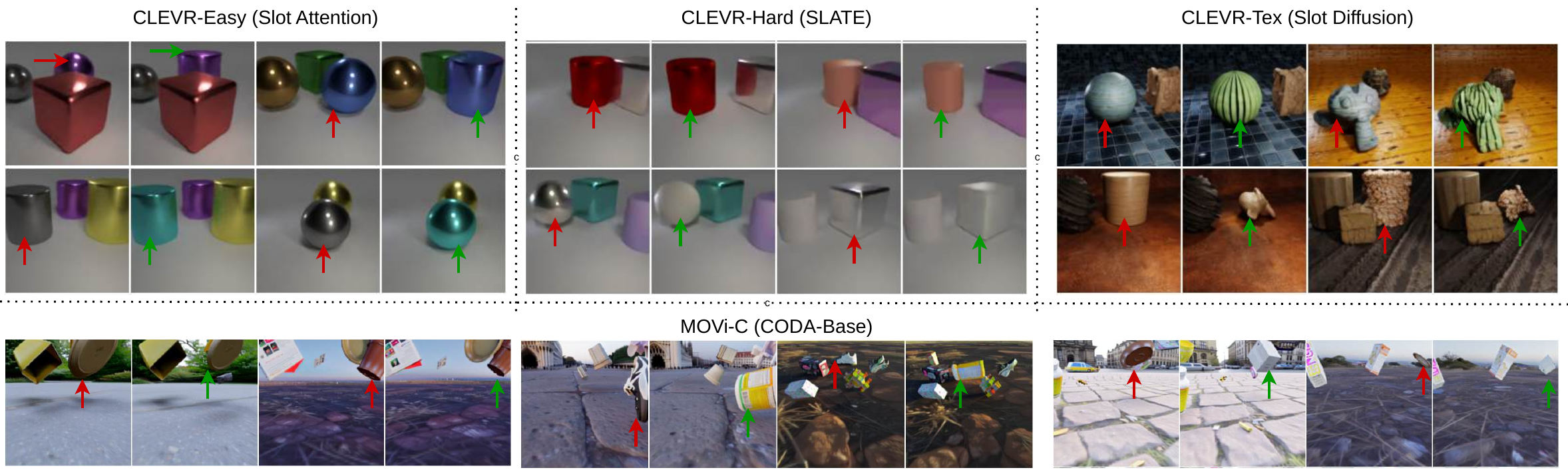} 
    \caption{\textbf{Object Manipulation via delta directions:} Delta direction is estimated by subtracting slot of object indicated from first image from slot of the object indicated in second image. This delta direction is then added to the object indicated in the third image. The fourth image is decoded by slots manipulated slots from third image. First three images are reconstruction. }
    \label{fig:Object_manuplation}
\end{figure*}

\begin{table}[t]
\centering
\small
\setlength{\tabcolsep}{3pt}
\renewcommand{\arraystretch}{1.1}
\caption{Cosine similarity between two estimations of the same delta direction.}
\label{tab:lrh_cosine_sim}

\begin{tabular}{lccccccc}
\toprule
& \multicolumn{2}{c}{CLEVR Easy}
& \multicolumn{2}{c}{CLEVR Hard}
& \multicolumn{2}{c}{CLEVR-Tex}
& MOVi-C \\
\cmidrule(lr){2-3}
\cmidrule(lr){4-5}
\cmidrule(lr){6-7}
\cmidrule(lr){8-8}

\textbf{Method}
& \textbf{Slot Attn.} & \textbf{SLATE+}
& \textbf{Slot Attn.} & \textbf{SLATE+}
& \textbf{SLATE+} & \textbf{Slot Diff.}
& \textbf{CODA} \\
\midrule

Same $\delta$
& 0.6610 & 0.7465
& 0.6024 & 0.6511
& 0.5618 & 0.6356
& 0.2230 \\

Different $\delta$
& 0.2232 & 0.1647
& 0.1898 & 0.1493
& 0.1498 & 0.1719
& 0.1020 \\

\bottomrule
\end{tabular}
\end{table}

\subsection{Linearity in Slots} \label{subsec:lrh_results}

As mentioned above, the presence of linear structure in the slot space implies the existence of \textit{global delta directions} \ref{note:linear_structure}. We test this by measuring the cosine similarity between two estimates of the same delta direction obtained from different object slots in two different images. For example, for the color factor, we compare $\delta_{\mathrm{color}}[\mathrm{red}\rightarrow\mathrm{blue}]$ estimated from two different object slots belonging to two different images. As a baseline, we measure the cosine similarity between delta directions corresponding to different factors. Table~\ref{tab:lrh_cosine_sim} shows the results across four datasets and four different methods. The cosine similarity between two estimates of the same delta direction (first row) is consistently higher than that between delta directions from different factors. This trend holds across all slot methods, including those using pretrained diffusion decoders, and across all datasets. These results suggest the existence of global delta directions and, consequently, linearity in the slot space, which is inherent to slot attention rather than specific to a particular dataset or model. On MOVi-C, CODA exhibits lower absolute values, which we attribute to errors in slot inference due to over- or under-segmentation.

We also manipulate slots using these delta directions, analogous to word-embedding arithmetic, where $r(\mathrm{king}) - r(\mathrm{boy})+ r(\mathrm{girl}) \sim r(\mathrm{queen})$. Specifically, we modify a slot by adding a delta direction and decode it using the slot decoder to generate the manipulated image. Figure~\ref{fig:Object_manuplation} shows the results. Changing a slot by adding a delta direction consistently changes the corresponding object factor in a predictable manner. This behavior holds from the weaker SBD slot decoder to the strongest diffusion-based decoder, suggesting that linearity is fundamental to slot representations. Please refer to Appendix \ref{appendix:lineariy_in_slots} for additional results showing tracing along the principal axes of delta directions. 
\subsection{Object and Factor Discovery}
\begin{figure*}[t] 
    \centering    
    \includegraphics[width=\textwidth]{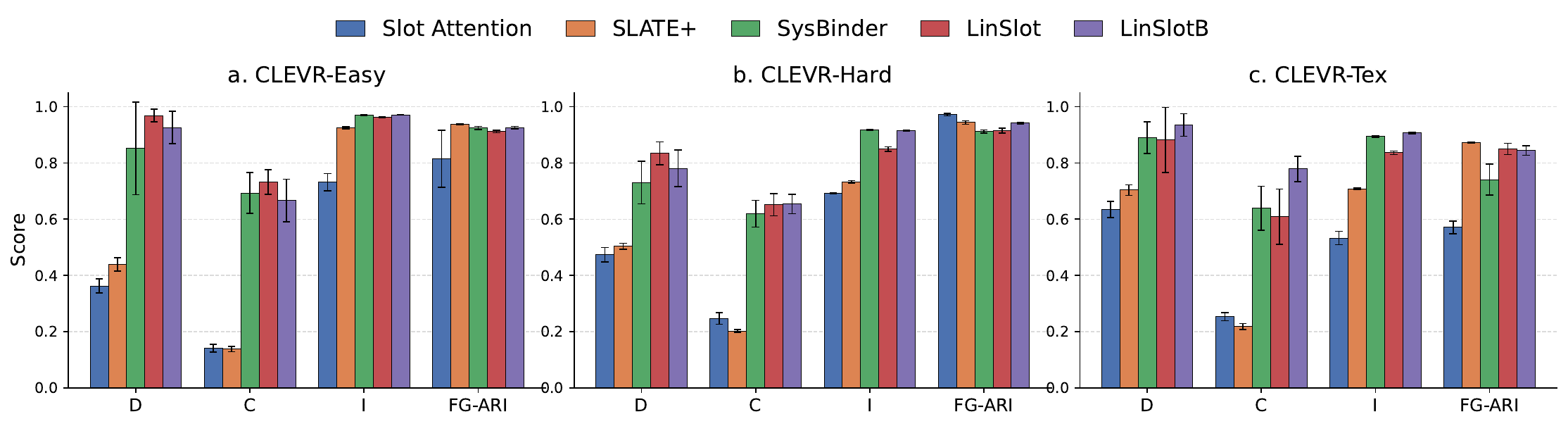} 
    \caption{\textbf{Quantitative Results:} Comparison of DCI and FG-ARI scores for our model and baselines across all datasets.}
    \label{fig:image_dci}
\end{figure*} \label{sec:quantitative_results_images}

\begin{figure*}[t]
    \centering
\includegraphics[width=\textwidth]{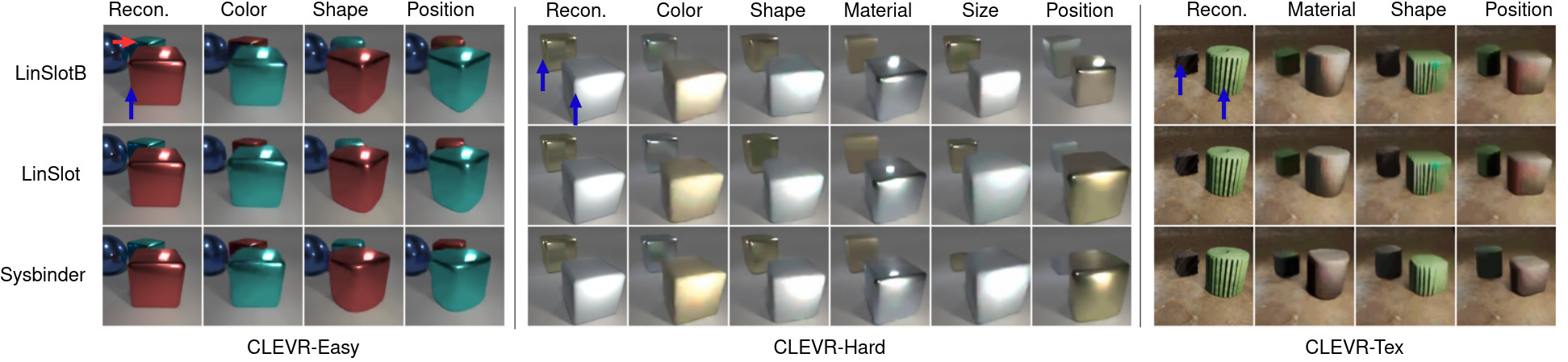} 
    \caption{\textbf{Qualitative Results:} Comparison of attribute swapping results between our model and SysBinder. Blue arrows indicate the objects participating in the swaps. \textit{Recon.} denotes the image reconstructed by the model.
    }
    \label{fig:image_swap}
\end{figure*}

Fig.~\ref{fig:image_dci} shows the DCI and FG-ARI scores for all the methods across all the datasets. Our attention-based methods with the slot decoder, \model, and the block decoder, \model B, perform better on the CLEVR-Easy and CLEVR-Hard datasets than all the baselines on the D, C, and FG-ARI scores. On CLEVR-Tex, \model \textbf{ }is on par with the best method on the D and C scores, whereas its FG-ARI score is comparable to the best number, which is achieved by \model B. This highlights the effectiveness of our approach, which does not require specialized decoders or the extensive hyperparameter tuning required by block-slot attention. The I score is lower than the best-performing method on the CLEVR-Hard and CLEVR-Tex datasets. However, when changing the decoder to the block decoder, as in \model B, the I scores match the best-performing values, suggesting the role of the decoder in the I score. The variance of both our methods is lower than that of the other baselines, suggesting lower sensitivity to the random seed. This suggests that our identifiability analysis is effective.

Additionally, we perform a swapping experiment in which we swap the blocks between two objects from the same image and decode the swapped representations to see whether the factors change in the intended way. Fig.~\ref{fig:image_swap} shows the results of both our methods along with SysBinder on all three datasets using their best seeds. We observe that on CLEVR-Easy, all methods perform the swapping faithfully. On CLEVR-Hard, we observe that even the best seed of SysBinder struggles with the size factor. On CLEVR-Tex, however, we observe that all the methods perform the swapping faithfully.

\subsection{Ablations}
 
\textbf{Effect of Training Choices:} 
Our method, \model, is trained in two phases and uses both the slots generated from the inferred blocks and the slots inferred directly from the images for decoding. We ablate the impact of these training choices, as they are not trivial consequences of the generative model. The results are shown in Table~\ref{tab:training_choices_ablation}. Removing either choice substantially affects the D and C scores, whereas removing the two-phase training and using slots inferred directly from the images leads to training collapse.


\textbf{GRU-MLP Sensitivity: } 
We use a GRU and MLP combination in the refinement loop of the blocks. This choice was inspired by Block-Slot Attention \cite{nsb}. We ablate these design choices; the results are shown in Table~\ref{tab:gru_mlp_ablation}. We note that \model\ is not sensitive to the choice of GRU and MLP, unlike Block-Slot Attention, highlighting robustness of our model.



\begin{table}[t]
\vskip 0.1in
\begin{small}
\begin{sc}
\begin{center}

\begin{minipage}{0.48\linewidth}
\centering
\caption{Effect of Training Choices}
\label{tab:training_choices_ablation}
\vspace{2pt}
\resizebox{\linewidth}{!}{%
\begin{tabular}{lccccc}
\toprule
2-Phase & Both & D & C & I & FG-ARI\\
\midrule
\textcolor{green}{\checkmark} & \textcolor{green}{\checkmark} & 0.9686 & 0.7322 & 0.9621 & 0.9122\\
\textcolor{red}{\texttimes} & \textcolor{green}{\checkmark} & 0.8466 & 0.5994 & 0.9487 & 0.9055\\
\textcolor{green}{\checkmark} & \textcolor{red}{\texttimes} & ----- & ----- & ----- & -----\\
\textcolor{red}{\texttimes} & \textcolor{red}{\texttimes} & 0.7592 & 0.6104 & 0.9368 & 0.9307\\
\bottomrule
\end{tabular}%
}
\end{minipage}
\hfill
\begin{minipage}{0.48\linewidth}
\centering
\caption{GRU-MLP Sensitivity}
\label{tab:gru_mlp_ablation}
\vspace{2pt}
\resizebox{\linewidth}{!}{%
\begin{tabular}{lccccc}
\toprule
GRU & MLP & D & C & I & FG-ARI\\
\midrule
\textcolor{green}{\checkmark} & \textcolor{green}{\checkmark} & 0.9686 & 0.7322 & 0.9621 & 0.9122\\
\textcolor{red}{\texttimes} & \textcolor{green}{\checkmark} & 0.9807 & 0.7322 & 0.9634 & 0.9113\\
\textcolor{green}{\checkmark} & \textcolor{red}{\texttimes} & 0.9710 & 0.7356 & 0.9619 & 0.9158\\
\textcolor{red}{\texttimes} & \textcolor{red}{\texttimes} & 0.9767 & 0.7468 & 0.9656 & 0.9189\\
\bottomrule
\end{tabular}%
}
\end{minipage}

\end{center}
\end{sc}
\end{small}
\end{table}

\section{Conclusion and Limitation}

In this work, we presented a generative framework for unsupervised object and factor discovery that exploits the linear additive structure of factors in slot representations. To the best of our knowledge, we are the first to demonstrate the Linear Representation Hypothesis (LRH) in slot space. We further proposed a sequential inference framework and derived its learning objectives from ELBO optimization. Experiments across datasets and slot-based methods provide consistent evidence for linear structure in slot representations, while our attention-based block inference effectively discovers both objects and their factors. Our model does not explicitly represent background or global properties such as lighting and style, which are treated as sources of uncertainty. Future work includes end-to-end identifiability, extension to video, and explicit modeling of global properties and object interactions.


\section{AI use statement}
In this work, we used generative AI tools for grammar and punctuation refining of text. We have not used generative AI tools for creating ideas, writing core code of the model, writing proofs. Additionally, we used generative AI tools for creating plot code. We have reviewed all AI-assisted work. We verified the grammar and graph generated. We take responsibility for the final content of this work, including text, claims or artifacts produced with the aid of generative AI.
\bibliography{iclr2027_conference}
\bibliographystyle{iclr2027_conference}

\appendix
\section{Theoretical Proofs}

\subsection{Proof of identifiability theorem} \label{appendix:proof_of_identifiablity}

Dropping the slot index for simplicity, and letting $c$ denote the concatenated factor vector, the generative model for a slot can be written as:
\begin{equation}
    s = f(\mathbf{c}) + \epsilon \quad \text{where} \quad \epsilon \sim \mathcal{N}(\mathbf{0}, \Sigma_S)
\end{equation}
Where $f(\mathbf{c}) = \sum_{i=1}^M f_{i}(c_i)$ is also a piecewise affine function. 
The distribution of the concatenated vector $c$ is a joint GMM:
\begin{equation}
    \mathbf{c} \sim \sum_{k=1}^K \bar{\pi}_k \mathcal{N}(\bar{\mu}_k, \bar{\Sigma}_k)
\end{equation}
where the total number of components is $K = \prod_{i=1}^{M} K_i$. Here, $\bar{\mu}_k$ and $\bar{\Sigma}_k$ are formed by concatenating the corresponding component means $\mu_{k,i}$ and block-diagonalizing the covariances $\Sigma_{k,i}$ across all factors $i$, and $\bar{\pi}_k$ is the product of the relevant mixing weights $\pi_{i,k}$.

According to the theorem established by \cite{identifiability_gmm}, the latent variable $c$ in this linear Gaussian model is identifiable up to permutation, scaling, and translation. Suppose we have two trained models that yield the identical marginal distribution for $s$:
\begin{align}
    s &\sim \mathcal{N}(f(\mathbf{c}), \Sigma) &&\text{where} \quad c \sim \sum_{k=1}^K \bar{\pi}_k \mathcal{N}(\bar{\mu}_k, \bar{\Sigma}_k) \\
    s &\sim \mathcal{N}(f'(\textbf{c'}), \Sigma') &&\text{where} \quad c' \sim \sum_{k=1}^K \bar{\pi}'_k \mathcal{N}(\bar{\mu}'_k, \bar{\Sigma}'_k)
\end{align}

Identifiability implies that the latent variables from the two models are related by:
\begin{equation}
    c = \Pi D c' + b
\end{equation}
where $\Pi$ is a permutation matrix, $D$ is a diagonal scaling matrix, and $b$ is a translation vector. 

We now argue that $\Pi$ must be a block-permutation matrix; that is, it cannot distribute the dimensions of a single factor vector $c'^u$ from the second model across two different factors $c^i$ and $c^v$ in the first model. 

Assume, for the sake of contradiction, that $\Pi$ does split an factor. Without loss of generality, suppose a subset of dimensions from $c'^u$ is assigned to $c^i$, and the remaining dimensions of $c'^u$ are assigned to a different factor $c^v$. 

Conversely, this means the factor $c^i$ in the first model is composed of scaled, permuted, and translated dimensions sourced from at least two independent factors in the second model (e.g., $c'^u$ and $c'^w$). Let these source factors have $K_u$ and $K_w$ mixture components, respectively. 

Intuitively, combining dimensions from different independent factors necessitates a multiplicative expansion in the number of mixture components to properly model the new joint space (i.e., it requires $K_u \times K_w$ components). However, our target factor $c^i$ is strictly defined as a sum over only $K_i$ components. Thus, mixing these dimensions would demand significantly more underlying components than the model's structural capacity allows.

Formally, because the source factors are independent, concatenating them forms a joint GMM with $K_u K_w$ components:
\begin{equation}
    c^{i} \sim \sum_{l=1}^{K_u K_w} \pi''_l \mathcal{N}(\mu''_l, \Sigma_l'')
\end{equation}
However, by our initial definition, $c^{i}$ itself is drawn from a GMM with exactly $K_i$ components:
\begin{equation}
    c^{i} \sim \sum_{l=1}^{K_i} \pi_{i,l} \mathcal{N}(\mu_{i,l}, \Sigma_{i,l})
\end{equation}
Equating these two distributions yields:
\begin{equation}
    \sum_{l=1}^{K_i} \pi_{i,l} \mathcal{N}(\mu_{i,l}, \Sigma_{i,l}) \equiv \sum_{l=1}^{K_u K_w} \pi''_{l} \mathcal{N}(\mu''_l, \Sigma_l'')
\end{equation}
This equivalence requires $K_i = K_u K_w$. However, we explicitly assumed that $K_i \neq K_u K_w$ for any triplet of distinct indices $(i, u, w)$. This is a contradiction. Therefore, $\Pi$ cannot split factors across different blocks, proving that $\Pi$ must be block-diagonal.

\subsection{Derivation of the ELBO} \label{sec:elbo_derivation}
Note that the generative process of the data is:
\begin{equation}
    p_{\theta}(X,\mathbf{S},\mathbf{C}) = p_{\theta_X}(X|\textbf{S})\prod_{i=1}^N \left ( p_{\theta_S}(s_i | \textbf{c}_i) \prod_{j=1}^M p_{\theta_{C_j}}(c_{j,i}) \right ) 
\end{equation}
We further define following joint two distributions
\begin{equation}
    p_{\theta_S}(\mathbf{S} | \mathbf{C}) = \prod_{i=1}^N p_{\theta_S}(s_i | \mathbf{c}_i) \ \ \ \text{and}  \ \ \ \ p_{\theta_C}( \mathbf{C}) = \prod_{i=1}^N \left (\prod_{j=1}^M p_{\theta_{C_j}}(c_{j,i})   \right )
\end{equation}

Further we also define the following joint distributions from variational posterior as:
\begin{equation}
    q_{\phi_C}(\textbf{C} | \textbf{S}) = \prod_{i=1}^N q_{\phi_C}(\textbf{c}_i | s_i)
\end{equation}

The log likelihood of the data can be written as:
\begin{align}
    \log p_{\theta}(X)
    &= \log \left ( \int_{\textbf{S}}\int_\textbf{C}
    p_{\theta_X}(X | \mathbf{S})
    p_{\theta_S}(\mathbf{S} | \mathbf{C})
    p_{\theta_C}(\mathbf{C})
    d\mathbf{S}d\mathbf{C}\right ) \\
    &\geq \mathbb{E}_{q_{\phi}(\textbf{S},\textbf{C}|X)}
    \left [ \log\left(
    \frac{p_{\theta_X}(X | \mathbf{S})
    p_{\theta_S}(\mathbf{S} | \mathbf{C})
    p_{\theta_C}(\mathbf{C})}
    {q_{\phi_S}(\textbf{S} | X)
    q_{\phi_C}(\textbf{C} | \textbf{S})}
    \right ) \right]\\
    &= \mathbb{E}_{q_{\phi_S}(\textbf{S} | X)}
    \left [ \log p_{\theta_X}(X | \textbf{S}) \right]
    - \underbrace{ D_{KL}\left(q_{\phi_S}(\textbf{S}|X)
    \,\|\,p_{\theta_S}(\mathbf{S}|\mathbf{C})\right)}_{\text{slot regularizer}} \\
    &\qquad
    - \underbrace{\mathbb{E}_{q_{\phi_S}(\textbf{S}|X)}
    \left[
    D_{KL}\left(q_{\phi_C}(\mathbf{C}|\mathbf{S})
    \,\|\,p_{\theta_C}(\mathbf{C})\right)
    \right]}_{\text{block regularizer}}.
\end{align}

\subsection{Concept Memory projection as Explicit block regularizer} \label{sec:concept_projection_elbo}
The block regularizer encourages the inferred block distribution to match the GMM prior. This regularizer can be factored at the object level and further decomposed into a sum of regularizers over individual blocks. We then use the result from \cite{gmm-vae-sample, gmm-vae} for the KL divergence between a normal distribution and a GMM:
 
\begin{align}
    D_{KL}\left(q_{\phi_C}(\mathbf{C}|\mathbf{S})
    \,\|\,p_{\theta_C}(\mathbf{C})\right) &= \sum _{i=1}^ND_{KL}\left(q_{\phi_C}(\mathbf{c}_i|s_i)
    \,\|\,p_{\theta_C}(\mathbf{c}_i)\right) \\ 
    &=  \sum _{i=1}^N \sum_{j=1}^M D_{KL}\left(q_{\phi_C}(c_{j,i}|s_i)
    \,\|\,p_{\theta_{C_j}}(c_{j,i})\right) 
\end{align}
Assuming that $q_{\phi_C}(c_{j,i}|s_i)$ is a normal distribution, each term in the above sum corresponds to the KL divergence between a normal distribution and a GMM. Using Jensen's inequality, this KL divergence can be upper-bounded by a weighted sum of KL divergences between the inferred distribution and the individual Gaussian components of the GMM.
\begin{align}
     D_{KL}\left(q_{\phi_C}(c_{j,i}|s_i)
    \,\|\,p_{\theta_{C_j}}(c_{j,i})\right) &= -\mathbb{E}_{q_{\phi_C}(c_{j,i}|s_i)} \left [ \log\left (  \frac{p_{\theta_{C_j}}(c_{j,i})}{q_{\phi_C}(c_{j,i}|s_i)} \right )\right] \\
    &=  -\mathbb{E}_{q_{\phi_C}(c_{j,i}|s_i)} \left [ \log\left (  \frac{\sum_{k=1}^{K_j} \pi_{k,j} \mathcal{N}(c_{j,i};\mu_{k,j}, \Sigma_{k,j})}{q_{\phi_C}(c_{j,i}|s_i)} \right )\right] \\
\end{align}
Introducing an auxiliary assignment distribution
$\gamma_{k,j}(s_i)$ satisfying
$\sum_k \gamma_{k,j}(s_i)=1$, the above expression can be upper-bounded as:
\begin{align}
     D_{KL}\left(q_{\phi_C}(c_{j,i}|s_i)
    \,\|\,p_{\theta_{C_j}}(c_{j,i})\right) &
    =  -\mathbb{E}_{q_{\phi_C}(c_{j,i}|s_i)} \left [ \log\left (  \frac{\sum_{k=1}^{K_j} \pi_{k,j} \mathcal{N}(c_{j,i};\mu_{k,j}, \Sigma_{k,j})\gamma_{k,j}(s_i)}{\gamma_{k,j}(s_i)q_{\phi_C}(c_{j,i}|s_i)} \right )\right] \\
    &\leq  \gamma_{k,j}(s_i) D_{KL}(  
    q_{\phi_C}(c_{j,i}|s_i) || \mathcal{N}(\mu_{k,j}, \Sigma_{k,j})) \\  
    &\qquad + \sum_{k}^{K_j} D_{KL} ( \gamma_{k,j}(s_i) || \pi_{k,j} )
\end{align}

Using the assumption that the $q_{\phi_C}(c_{j,i} | s_i)$ is normal distribution with mean $c_{j,i}$ and additionally that it have identity covariance matrix. Also assuming all the $\Sigma_{k,j}$ are also identity. Also ignoring the last term then the the best $\tilde{c}_{j,i}$ will be when the KL divergence is minimized.
\begin{equation}
   c_{j,i}^*  = \min_{c_{j,i}} \sum_{k=1}^{K_j} \gamma_{k,j}(s_i)||c_{j,i} - \mu_{k,j}||_2^2
\end{equation}
Which have the solution that 
\begin{equation}
    c_{j,i}^* = \sum_{k=1}^{K_j} \gamma_{k,j}(s_i) \mu_{k,j}
\end{equation}

Thus, the optimal block representation is a weighted combination of the GMM prototype vectors. In particular, if the assignment weights are given by
$\gamma_{k,j}(s_i)=\frac{\exp({\texttt{cosine-sim}(c_{j,i},\mu_{k,j}))}}{\sum_{k}\exp({\texttt{cosine-sim}(c_{j,i},\mu_{k,j}))}}$
and the prototype matrix is defined as
$\mathcal{M}_j^T = \begin{bmatrix}
    \mu_{1,j} \ \  
    ..   \ \ \mu_{K_j,j}
\end{bmatrix}$
then the resulting representation corresponds to a projection of the block onto the prototype space.

This establishes the connection between the concept memory projection and the explicit block regularizer.

\section{Experiment Details} \label{appendix:training}
\subsection{Two Phase training}\label{appendix:two_phase_training}

Please refer to fig \ref{fig:training} where we show our two phase training pipeline. In phase one we train the slot attention and decoder on the slot method specific loss. In phase two we add the block specific modules and keep the decoder frozen.  

\begin{table}[t]
    \caption{Hyperparameters of training for CLEVR-Easy, CLEVR-Hard, and CLEVR-Tex datasets.}
    \label{tab:hyperparameters_clevr}
    \vskip 0.1in
    \centering
    \setlength{\tabcolsep}{1mm}
    \begin{tabular}{l|l|c|c|c}
    \hline
    Module & Hyperparameter & \multicolumn{3}{c}{Dataset} \\
    & & CLEVR & CLEVR & CLEVR\\
     & & Easy & Hard & Tex \\
    \hline
    General & Batch Size & 64 & 64  & 64 \\
    & Image Size & 128 & 128 & 128 \\
    & Training Steps & 200K & 260K  & 315K  \\
    & Encoder Learning Rate & 0.0001 & 0.0001 &  0.0001\\
    & Decoder Learning Rate (only used in phase 1) & 0.0003 & 0.0003 &  0.0003\\
    \hline
    Block & Block Size ($d_{block}$) & 192 & 192 & 192  \\
    Refinement & \# Blocks ($M$) & 8 & 16  &  8 \\
    & \# Concepts ($K_j$ same across all $j$) & 64 & 64 & 64 \\
    & \# Refinement Steps ($T_b$) & 3 & 3 & 3 \\
    & \# Projections ($N_p$) & 192 & 192  & 192 \\ 
    \hline
    Slot & Slot Size ($d_{slot}$) & 192 & 192 & 192  \\
    Attention & \# Slots ($N$) & 4 & 4  &  6 \\
    & \# Refinement steps ($T_s$) & 3 & 3 & 3 \\
    \hline
    \end{tabular}
\end{table}

 \begin{figure}[t]
    \centering
        \includegraphics[width=0.8\linewidth]{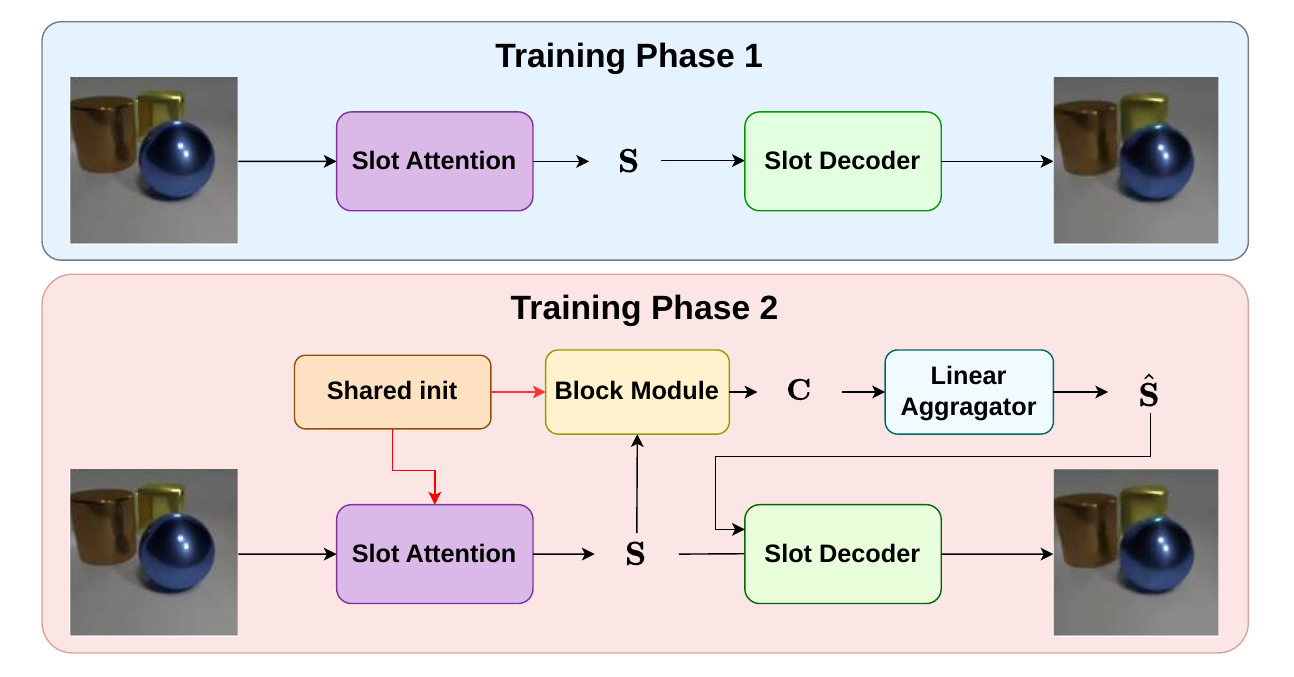}
        \label{fig:training}
    \caption{Training phases of the model.}
    \label{fig:training}
\end{figure}

\subsection{Additional Details of the datasets}
\label{appendix:datasets}
We use three image benchmarks derived from \cite{Johnson2016CLEVRAD, clevrtex}, provided by \cite{nsb}. These datasets consist of the multi object images with $128 \times128$ resolution. The objects in these datasets are characterized by set of categorical and continuous attributes with details as below. Additionally for slot linearity validation we further use MOVi-C \cite{movi-c}

\textbf{\underline{CLEVR-Easy}} dataset features scenes with 2-3 objects. Each object have 3 attributes: color with 8 possible choices, shape with 3 possible choices and position which is continuous.

\textbf{\underline{CLEVR-Hard}}, a more challenging dataset with images consisting of 2-3 objects. Each object has 5 attributes: color with 137, shape with 3 and material with 2 possible choices; size comes from fixed range and the position is continuous. 

\textbf{\underline{CLEVR-Tex dataset}} consists of textured images where the objects and background both have realistic textures increasing the visual complexity. The images consist of 2-3 objects with each object having 3 attributes. Texture/Material have 57 possible choices, shapes have 4 possible choices and position is continuous.

\textbf{\underline{MOVi-C}} is the originally video dataset of objects rendered on the natural background. The frames from video as used as images.

\subsection{Creation of the paired Dataset}\label{appendix:paired_dataset}
Each datapoint of the paired dataset is a tuple $t^{(l)}=(X^{(l,1)}, X^{(l,2)}, X^{(l,3)}, X^{(l,4)})$
of four images. The images $X^{(l,1)}$ and $X^{(l,2)}$ contain the same objects except one;
without loss of generality, let it be the first object generated in both images. These two
objects differ in exactly one attribute $j^\star \in \{1,\dots,M\}$, i.e.
$c_{j,1}^{(l,1)} = c_{j,1}^{(l,2)}$ for all $j \in \{1,\dots,M\}\setminus\{j^\star\}$.
Similarly, $X^{(l,3)}$ and $X^{(l,4)}$ contain the same objects except the first one, and
these first objects differ only in attribute $j^\star$:
$c_{j,1}^{(l,3)} = c_{j,1}^{(l,4)}$ for all $j \in \{1,\dots,M\}\setminus\{j^\star\}$.
Moreover, both pairs undergo the same change of attribute $j^\star$:
$c^{(l,1)}_{j^\star,1} = c^{(l,3)}_{j^\star,1}$ and $c^{(l,2)}_{j^\star,1} = c^{(l,4)}_{j^\star,1}$,
while the first objects of the two pairs differ in at least one other attribute,
$\exists\, j \neq j^\star : c^{(l,1)}_{j,1} \neq c^{(l,3)}_{j,1}$.

\section{Additional Results} \label{appendix:results}
\subsection{Slate modifications}\label{appendix:Slate_modifications}

We modified the original SLATE \cite{slate} with initializing the slots as linear transformation of 2048 dimensional random vector sampled from learned gaussian rather than from just the learned normal distribution.  Additionally we add a learnable linear projector on the slots before passing them to transformer slot decoder which makes 8 different projections of slots. The transformer decoder then uses this $8N$ tokens for decoding same as block decoder does. Refer table \ref{tab:slate_appendix_table} which shows comparison on the CLEVR-Easy dataset.

\begin{table}[t]
\caption{SLATE vs SLATE+ Results on CLEVR-Easy}
\label{tab:slate_appendix_table}
\vskip 0.1in
\begin{small}
    \begin{sc}
\begin{center}
\begin{tabular}{lc}
\toprule 
Method & FG-ARI\\
\midrule
SLATE & 0.8166\\
SLATE+ & 0.9369\\
\bottomrule
\end{tabular}
\end{center}
\end{sc}
\end{small}
\end{table}

\subsection{Linearity in Slots} \label{appendix:lineariy_in_slots}

\begin{table}[t]
\centering
\caption{Stable rank and participation ratio of the delta-direction subspace of each factor (mean $\pm$ std over seeds).}
\label{tab:delta_rank}
\small
\setlength{\tabcolsep}{5pt}
\begin{tabular}{llccc}
\toprule
Dataset (model) & Factor & \#Values & Stable rank & Part.\ ratio \\
\midrule
\multirow{3}{*}{\shortstack[l]{CLEVR-Easy\\(Slot Attention, $d_{\text{slot}}{=}64$)}}
 & Position & cont. & $3.21 \pm 0.38$ & $5.58 \pm 0.53$ \\
 & Color    & 8     & $2.93 \pm 0.62$ & $4.91 \pm 1.46$ \\
 & Shape    & 3     & $2.06 \pm 0.16$ & $3.70 \pm 0.48$ \\
\midrule
\multirow{5}{*}{\shortstack[l]{CLEVR-Hard\\(SLATE+, $d_{\text{slot}}{=}192$)}}
 & Position & cont. & $3.89 \pm 0.05$ & $8.08 \pm 0.04$ \\
 & Color    & 137   & $3.82 \pm 0.11$ & $7.24 \pm 0.34$ \\
 & Shape    & 3     & $2.79 \pm 0.19$ & $5.37 \pm 0.52$ \\
 & Material & 2     & $1.83 \pm 0.07$ & $3.18 \pm 0.21$ \\
 & Size     & cont. & $1.82 \pm 0.12$ & $3.12 \pm 0.34$ \\
\midrule
\multirow{3}{*}{\shortstack[l]{CLEVR-Tex\\(Slot Diffusion, $d_{\text{slot}}{=}192$)}}
 & Position & cont. & $3.71 \pm 0.17$ & $6.60 \pm 0.54$ \\
 & Material & 57    & $5.43 \pm 0.37$ & $11.12 \pm 0.21$ \\
 & Shape    & 4     & $2.81 \pm 0.17$ & $6.46 \pm 0.59$ \\
\bottomrule
\end{tabular}
\end{table}

\begin{figure}
    \centering
    \includegraphics[width=\linewidth]{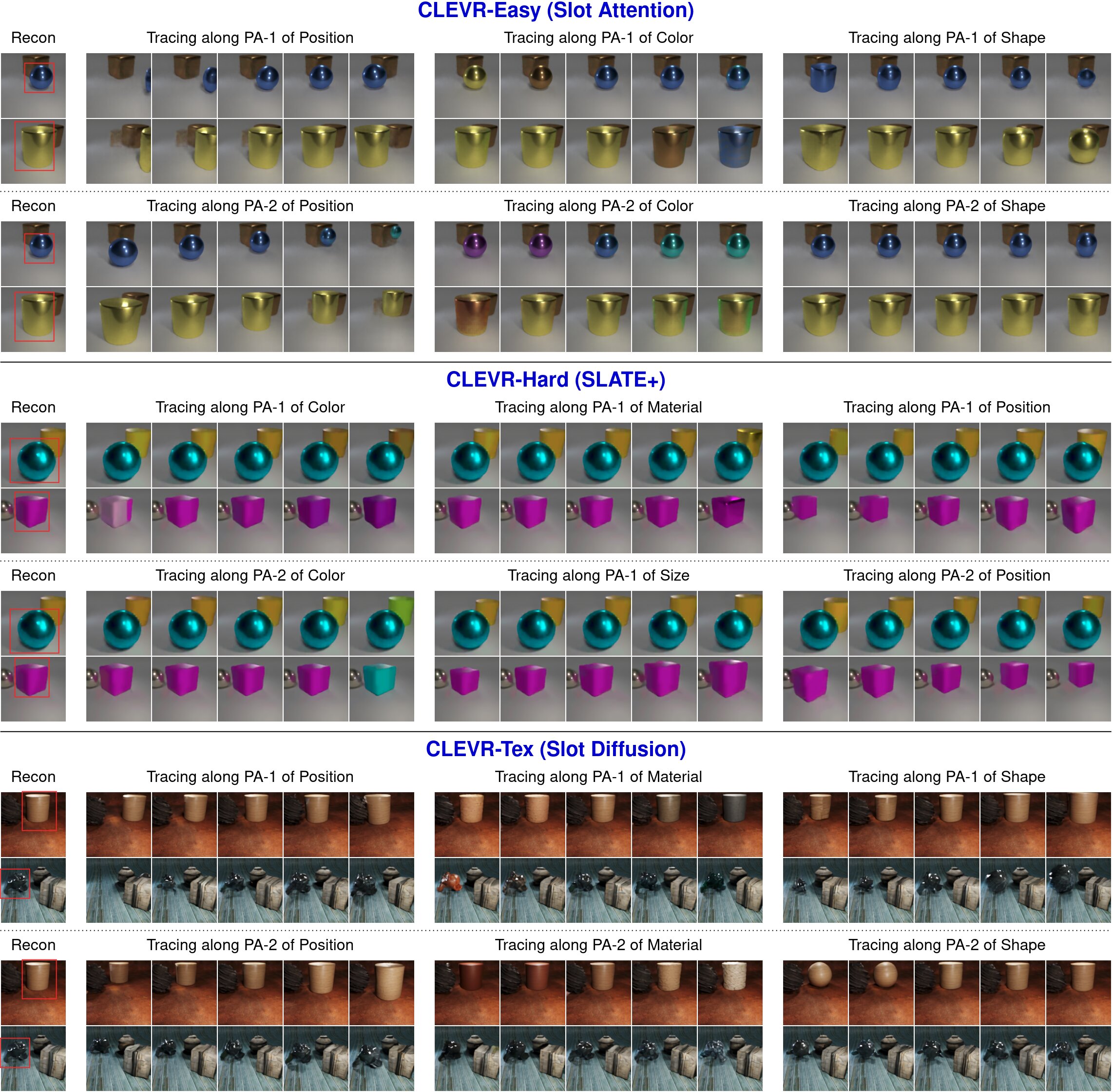}
    \caption{\textbf{Steering slots along principal axes of the delta space:}
    We steer slots by adding a scaled multiple $\alpha$ of a principal axis computed from the delta directions associated with a particular factor. The slots corresponding to the objects highlighted by red boxes in the first column are steered, with $\alpha$ increasing from left to right. The resulting images exhibit meaningful and smooth changes along the corresponding factor, consistently across different objects, datasets, and methods, including Slot Attention, SLATE+, and Slot Diffusion.}
    \label{fig:tracing_along_pa}
\end{figure}
The presence of the \textit{delta directions} in the slot space supports the linearity of the slot space. However, it does not reveal the structure of the factor manifold. One hypothesis is that the factor manifold is a \textit{linear subspace} of the slot space itself. The question is then what the inherent dimension of this subspace is and how to find its basis. 

Assuming the factor manifold is a linear subspace, the delta directions of a specific factor span this subspace. Finding the principal axes of the delta directions then naturally gives a basis for it. Specifically, we estimate the delta directions from all slot pairs in the paired data where factor $j$ is changed. We then center the directions by subtracting the mean delta direction and arrange them as the rows of a matrix $\Delta_j$. Next, we compute the singular value decomposition $\Delta_j = U\Sigma V^T$. We choose the top $l_j$ right singular vectors (the columns of $V$) as the principal axes, denoted as $\{p_{j,1},\dots,p_{j,l_j}\}$. Here, $l_j$ is the floor of the \textit{participation ratio} $\mathrm{PR}_j = \big(\sum_i \sigma_i^2\big)^2 / \sum_i \sigma_i^4$, where $\sigma_i$ are the singular values of $\Delta_j$. We then steer the slot $s$ of an object in an image as $s' = s + \alpha\, p_{j,i}$ for each principal axis $p_{j,i}$. We visualize the resulting changes by decoding the steered slot together with the remaining, unmodified slots. The range of $\alpha$ is set by the minimum and maximum projections of all slots in the dataset onto $p_{j,i}$. 

 Table~\ref{tab:delta_rank} reports the stable rank and participation ratio of the delta spaces across different slot methods and datasets. The delta spaces consistently have substantially lower effective dimensionality than the full slot space, indicating that changes associated with individual factors occupy low-dimensional directions within the slot representation. We also observe a relationship between the number of possible values of a factor and the dimensionality of its delta space, as measured by both stable rank and participation ratio, with material on CLEVR-Tex being a notable exception. We further examine whether these low-dimensional directions correspond to semantically meaningful variations by steering slots along the principal axes. Figure~\ref{fig:tracing_along_pa} shows representative results across datasets and methods. Steering along the principal axes consistently changes the corresponding factor while preserving the other properties of the object, and produces smooth transitions across different objects and datasets. This behavior is also consistent across random seeds. Moreover, individual principal axes often correspond to interpretable variations within a factor. For example, the first color principal axis on CLEVR-Hard primarily changes the brightness or darkness of the object color, while the two principal axes associated with position on CLEVR-Tex correspond approximately to horizontal and vertical image position. These results suggest that factor-specific variations are concentrated in low-dimensional, approximately linear subspaces of the slot space, and that the principal axes of these subspaces capture semantically coherent directions. Together, these findings provide empirical support for our hypothesis that factor manifolds are approximately linear in the learned slot space.

\end{document}